\documentclass[11pt,a4paper]{article}
\usepackage[OT1]{fontenc}
\usepackage[hyperref]{AACL-IJCNLP2020}
\usepackage{times}
\usepackage{latexsym}

\usepackage[hang,flushmargin]{footmisc}

\usepackage{url}
\usepackage{graphicx}
\usepackage{amsfonts}
\usepackage{amsmath}

\usepackage{microtype}

\aclfinalcopy 

\title{Investigating Label Bias in Beam Search for Open-ended Text Generation}

\author{Liang Wang$^1$, Jinlong Liu$^1$, Jingming Liu$^1$ \\
  $^1$Yuanfudao AI Lab, Beijing, China \\
  {\tt \{wangliang01,liujl09,liujm\}@fenbi.com}}

\date{}

\begin{document}
\maketitle
\begin{abstract}
Beam search is an effective and widely used decoding algorithm
in many sequence-to-sequence (seq2seq) text generation tasks.
However,
in open-ended text generation,
beam search is often found to produce repetitive and generic texts,
sampling-based decoding algorithms like top-k sampling and nucleus sampling are more preferred.
Standard seq2seq models suffer from label bias due to its locally normalized probability formulation.
This paper provides a series of empirical evidence
that label bias is a major reason
for such degenerate behaviors of beam search.
By combining locally normalized maximum likelihood estimation
and globally normalized sequence-level training,
label bias can be reduced with almost no sacrifice in perplexity.
To quantitatively measure label bias,
we test the model's ability to discriminate
the groundtruth text and a set of context-agnostic distractors.
We conduct experiments on large-scale response generation datasets.
Results show that beam search can produce more diverse and
meaningful texts with our approach,
in terms of both automatic and human evaluation metrics.
Our analysis also suggests several future working directions
towards the grand challenge of open-ended text generation.
\end{abstract}

\section{Introduction}
Neural text generation usually involves transforming some inputs into text outputs.
In directed generation ~\cite{Holtzman2019TheCC} tasks
including machine translation, summarization, and data-to-text, etc,
the output space is highly constrained by the given input.
Beam search ~\footnote{Unless explicitly specified, we use length normalization by default.}
is the de facto sequence decoding algorithm,
and provides good performance empirically.
By contrast,
in the more challenging open-ended text generation scenarios,
such as response generation and story generation,
there exist many plausible outputs given an input.
The outputs of beam search are often generic, repetitive, and meaningless.
Top-k sampling ~\cite{Radford2019LanguageMA}
and nucleus sampling (also referred to as top-p sampling) ~\cite{Holtzman2019TheCC}
are much more widely adopted.

\begin{figure}[ht]
\begin{center}
 \includegraphics[width=0.95\linewidth]{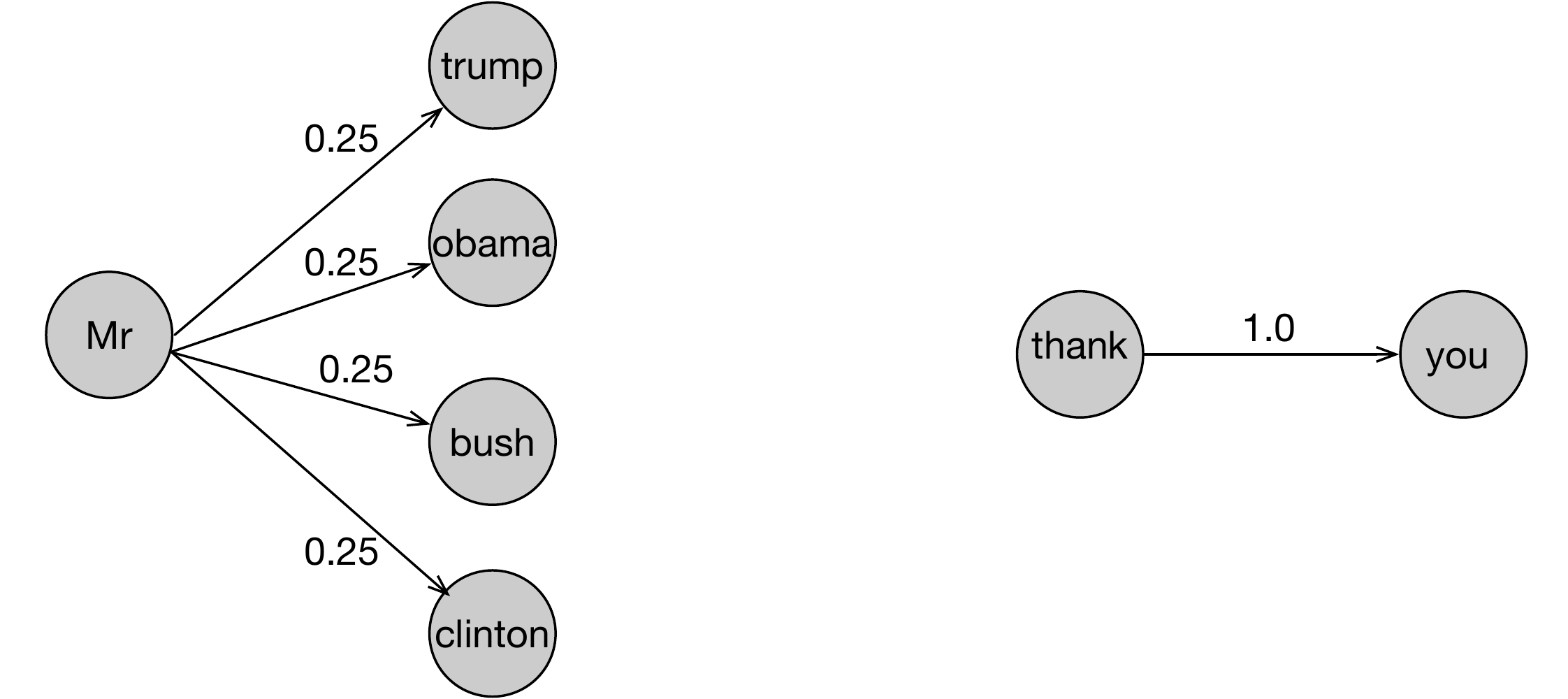}
 \caption{An illustrative example of label bias.
 Even though all hypotheses are plausible,
 the right one will be preferred because of the larger local probability.}
 \label{fig:label_bias}
\end{center}
\end{figure}

Label bias ~\cite{Hannun2020TheLB, Lafferty2001Conditional} refers to the phenomenon that
locally normalized models for structured prediction
often prefer output states with fewer outgoing transitions.
From the perspective of information theory,
models favor output states whose next-state distributions have low conditional entropy.
As shown in Figure ~\ref{fig:label_bias},
``Mr'' can be followed by many plausible words,
since the probability is locally normalized,
each word can only receive a small proportion of probability mass.
In the extreme case,
if the state transitions are deterministic,
the inputs will be completely ignored.
Label bias also makes it difficult to
correct past mistakes given new observations ~\cite{Lafferty2001Conditional}.

Seq2seq models with MLE factorize the probability of a sequence
into products of locally normalized probabilities.
As a result,
they also suffer from label bias.
Nonetheless,
it has been unclear whether
label bias has any connection with the degenerate behaviors of beam search or not.
Previous works ~\cite{Li2016ADO, Xu2019NeuralRG} propose some heuristic methods to mitigate this issue.
In this paper,
we evaluate the likelihood distribution of human texts and generated texts,
and show that beam search outputs are heavily biased towards the low-perplexity region.

To reduce label bias,
one can replace local normalization with globally normalized training.
In the seq2seq literature,
it is also called sequence-level training.
There are some successes in applying global normalization to
part-of-speech tagging ~\cite{Andor2016GloballyNT},
machine translation ~\cite{Edunov2018ClassicalSP}, etc.
Existing methods use average log-probabilities as the unnormalized score,
which are still based on local probabilities
and often severely hurt perplexity on held-out datasets.
In this paper,
we use unnormalized logits instead.
By combining token-level likelihood loss and sequence-level loss,
the logits can be calibrated
while keeping the local probabilities unchanged.

Evaluating open-ended text generation systems is non-trivial ~\cite{Liu2016HowNT}.
To verify the effectiveness of the proposed method,
it is important to be able to measure label bias.
We propose a heuristic ranking based metric.
First,
a set of low-perplexity texts are selected based on a pre-trained language model,
then these distractors and the groundtruth are ranked based on the predicted model scores.
A model with less label bias should rank the groundtruth before context-agnostic distractors.

We conduct experiments on a large-scale response generation dataset ConvAI2 ~\cite{Dinan2019TheSC}.
In terms of automatic evaluation metrics,
our method can produce significantly more diverse texts
than standard token-level MLE training
and has less negative impacts on perplexity.
Human evaluation gives a higher specificity and
sensibleness score ~\cite{Adiwardana2020TowardsAH} to our model's outputs.
Yet there is also some evidence showing
the label bias issue is still far from being solved.
We discuss several limitations of our work
and provide possible future directions to
better understand beam search in open-ended text generation.

\section{Likelihood Distribution Evaluation}

To better understand the degenerate behaviors of beam search,
we analyze the properties of perplexity distribution
for both human texts and generated texts.
Our analysis is based on the validation dataset of ConvAI2 ~\cite{Dinan2019TheSC}.
An off-the-shelf GPT2-117M model ~\footnote{\url{https://github.com/openai/gpt-2}}
is used to evaluate the unconditional language model perplexity.
We train a Transformer-based seq2seq model with standard MLE to evaluate the perplexity of responses.
Please check out Section ~\ref{section:setup} for more details about the dataset and our model.

\subsection{Beam Search Outputs are Heavily Biased}

Standard seq2seq models are trained under the principle of maximum likelihood:
maximize the probability of human texts given input contexts.
Naturally,
one would expect that
the texts generated by a well-trained model
should share similar characteristics with human texts.

\begin{figure}[ht]
\begin{center}
 \includegraphics[width=1.0\linewidth]{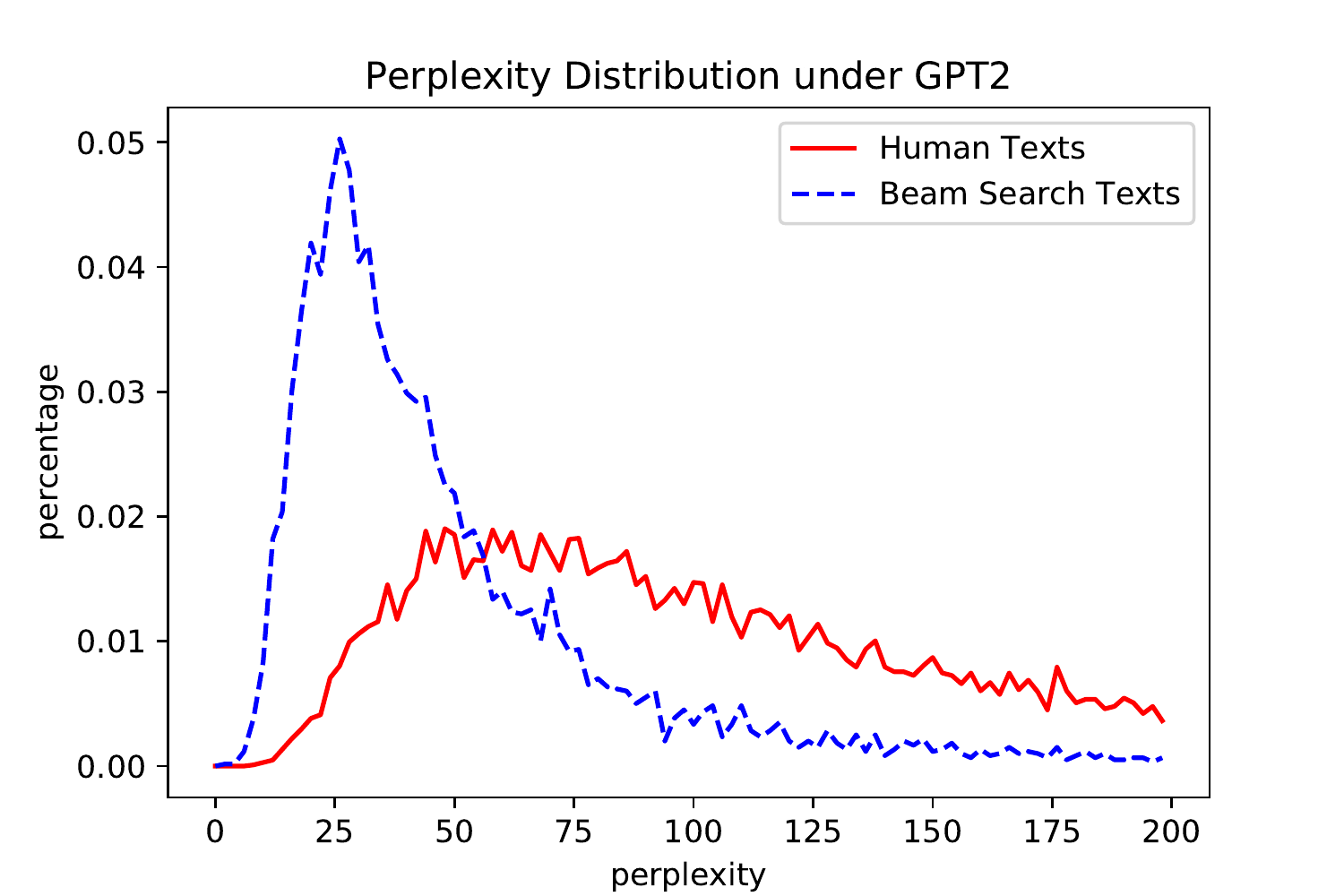}
 \caption{Perplexity distribution of human texts and generated texts under the pre-trained GPT2-117M language model.}
 \label{fig:ppl_gpt}
\end{center}
\end{figure}

In Figure ~\ref{fig:ppl_gpt},
we show the perplexity distribution under GPT2-117M.
The texts generated by beam search are heavily biased towards the low-perplexity region.
In contrast,
the distribution of human texts is flat
and has a long tail in the high-perplexity region.
Perplexity can be intuitively interpreted as the expected number of plausible next tokens.
Lower perplexity means beam search favors output states with fewer outgoing transitions,
which is a typical symptom of label bias.

\subsection{Search Errors or Model Errors?}

\begin{figure}[ht]
\begin{center}
 \includegraphics[width=1.0\linewidth]{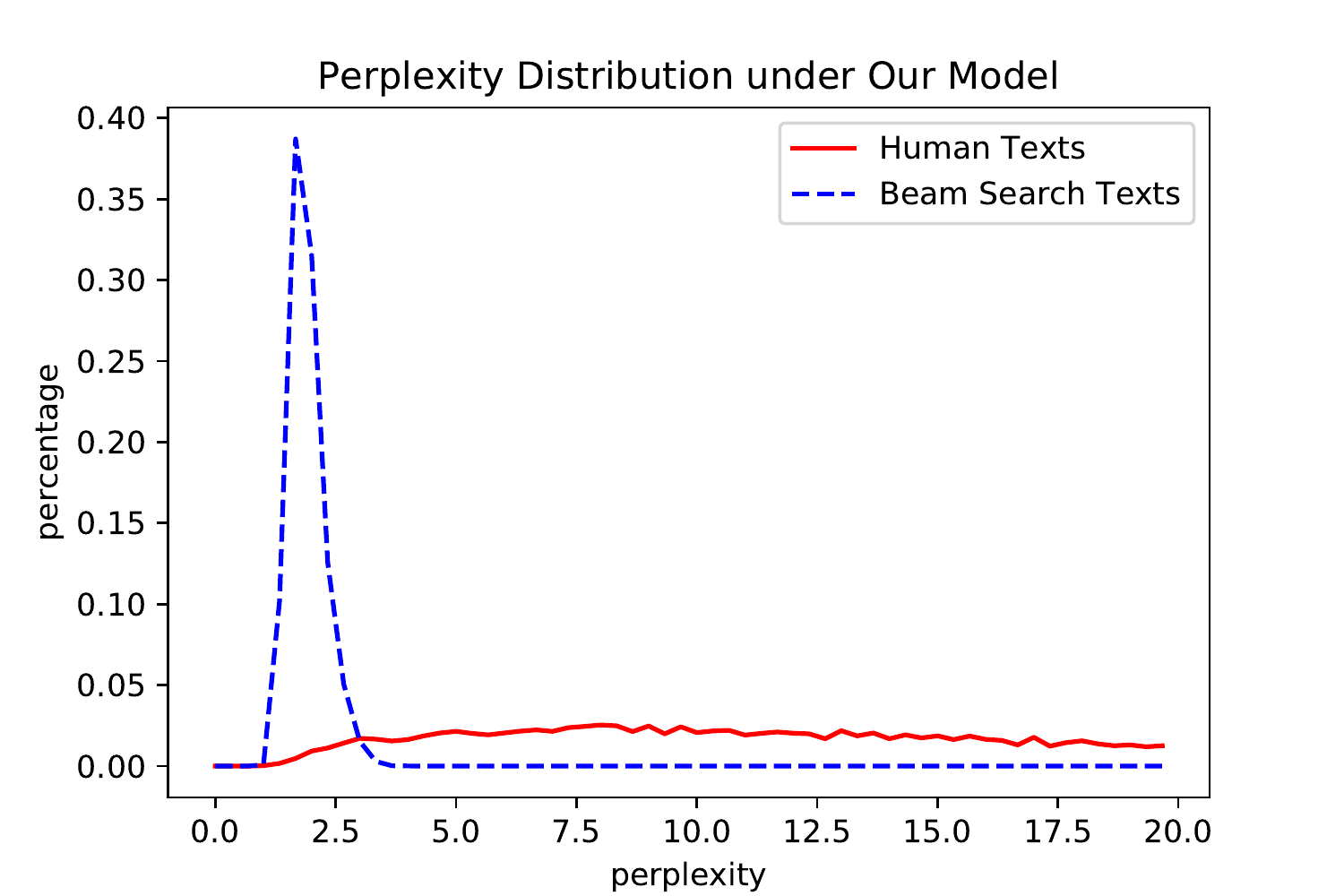}
 \caption{Perplexity distribution of human texts and generated texts under our trained response generation model.
 With access to input contexts,
 the perplexity is lower than the counterpart in Figure ~\ref{fig:ppl_gpt}.}
 \label{fig:ppl_model}
\end{center}
\end{figure}

There are two types of errors in seq2seq models using beam search decoder:
search errors and model errors.
For search errors,
a trained model assigns a higher score to the groundtruth text but beam search fails to find it.
For model errors,
beam search outputs indeed have higher scores under the model distribution.
With MLE training,
we use length-normalized log-probability as the score.

In Figure ~\ref{fig:ppl_model},
the average perplexity of generated texts is extremely low.
We can conclude empirically that
the degenerate behaviors of beam search are mainly
attributed to model errors instead of search errors.
Designing better search algorithms to find hypotheses
closer to the global optimum is unlikely to help.
A state-of-the-art conversational model Meena ~\cite{Adiwardana2020TowardsAH}
uses sample-then-rerank as decoding algorithm.
It first samples several candidates from the model distribution and
then reranks candidates based on perplexity.
Our findings imply that sample-then-rerank may risk producing degenerate outputs.

\section{Method}

\subsection{Token-Level Training and Inference} \label{section:token_loss}
Given input $\mathbf{x}=\{\mathbf{x}_i\}_{i=1}^{|\mathbf{x}|}$,
and target output $\mathbf{y}=\{\mathbf{y}_i\}_{i=1}^{|\mathbf{y}|}$,
seq2seq models aim to maximize the conditional probability $P(\mathbf{y}|\mathbf{x}, \mathbf{\Theta})$,
where $\mathbf{\Theta}$ denotes model parameters.
Standard token-level MLE training with teacher forcing factorizes this objective
in an auto-regressive way:
\begin{equation}
    P(\mathbf{y}|\mathbf{x}, \mathbf{\Theta}) = \Pi_{i=1}^{|\mathbf{y}|}P(\mathbf{y}_i|\mathbf{x}, \mathbf{y}_{<i})
\end{equation}

We omit $\mathbf{\Theta}$ to make the equation less cluttered.
Thus,
token-level cross entropy loss for an input-output pair $(\mathbf{x}, \mathbf{y})$
can be defined as follows:
\begin{equation}
    L_{token} = -\frac{1}{|\mathbf{y}|}\sum_{i=1}^{|\mathbf{y}|}\log P(\mathbf{y}_i|\mathbf{x}, \mathbf{y}_{<i})
\end{equation}

At inference stage,
given an input $\mathbf{x}$,
decoding algorithm attempts to search $\mathbf{\hat{y}}$
with highest average log-probability score:
\begin{equation} \label{equ:avg_logprob}
    \mathbf{\hat{y}} = \arg\max_{y'} \frac{1}{|\mathbf{y'}|}\sum_{i=1}^{|\mathbf{y'}|}\log P(\mathbf{y'}_i|\mathbf{x}, \mathbf{y'}_{<i})
\end{equation}

The search space grows exponentially to the output length,
heuristic algorithms like beam search are often used.
Beam search with beam size $k$ keeps $k$-best hypotheses at each time step.
The search procedure stops when there are $k$ complete hypotheses available
and it is impossible to get better hypotheses by expanding beams.

\subsection{Sequence-Level Training}
Token-level training suffers from label bias,
since the next-token probability are locally normalized over the vocabulary $\mathbf{V}$:
$\sum_{\mathbf{y'}_i \in \mathbf{V}}P(\mathbf{y'}_i|\mathbf{x}, \mathbf{y}_{<i}) = 1$.
For a specific timestep,
assume there are $n$ equally plausible tokens,
due to the constraint of local normalization,
each token will receive probability mass $\frac{1}{n}$.
Outputs with smaller $n$ have lower entropy for next-token distribution,
and will receive higher scores.
In open-ended text generation,
$n$ often varies in a large range.
Thus,
beam search will favor generic texts that
have smaller $n$ than human texts.

Sequence-level training explicitly maximizes the global score of groundtruth $y$.
There are several possible formulations,
such as empirical risk minimization,
margin-based sequence loss ~\cite{Edunov2018ClassicalSP}, etc.
Most formulations require an automatic metric such as BLEU
to calculate the score of a hypothesis,
which is not available in open-ended generation scenarios.
In this paper,
we cast sequence-level training as multi-class classification,
given a score function $s(\mathbf{y}, \mathbf{x})$,
sequence cross entropy loss is defined as:
\begin{equation} \label{equ:seq_loss}
    L_{seq} = -\log \frac{\exp(s(\mathbf{y}, \mathbf{x}))}{\sum_{\text{all}\ \mathbf{y'}}\exp(s(\mathbf{y'}, \mathbf{x}))}
\end{equation}

In Equation ~\ref{equ:seq_loss},
the denominator is often called ``partition function''
in the literature of Markov Random Fields ~\cite{Koller2009ProbabilisticGM},
the denominator is a sum over all possible output sequences.
The score function $s(\mathbf{y}, \mathbf{x})$ is learned by seq2seq models.
One common choice of score function is the average log-probability
as shown in Equation ~\ref{equ:avg_logprob}.
However,
this score function still builds upon the locally normalized probability
and often results in much worse perplexity.

Assume $u_{\mathbf{y}_i} \in (-\infty, +\infty)$ is
the unnormalized logit for token $\mathbf{y}_i$.
We propose to use the logits as scores:
\begin{equation} \label{equ:logits_score}
    s_{logit}(\mathbf{y}, \mathbf{x}) = \frac{1}{|\mathbf{y}|}\sum_{i=1}^{|\mathbf{y}|} u_{\mathbf{y}_i}
\end{equation}
One advantage is $u_{\mathbf{y}_i}$ can vary within a larger range than the log-probability.
Also,
for any real number $\Delta$,
\begin{equation} \label{equ:calibrate_logits}
\begin{aligned}
    P(\mathbf{y}_i) &= \frac{\exp(u_{\mathbf{y}_i})}{\sum_{j=1}^{|\mathbf{V}|}\exp(u_j)} \\
                    &= \frac{\exp(u_{\mathbf{y}_i} + \Delta)}{\sum_{j=1}^{|\mathbf{V}|}\exp(u_j+\Delta)}
\end{aligned}
\end{equation}
which indicates it is possible to calibrate the logits flexibly
by learning $\Delta$ without affecting $P(\mathbf{y}_i)$.

The final loss function $L$ is a linear combination of
token-level loss $L_{token}$ and sequence-level loss $L_{seq}$:
\begin{equation}
    L = \alpha L_{token} + \beta L_{seq}
\end{equation}
Empirically,
we set $\alpha=1$ and $\beta=5$.

\subsection{Partition Function Estimation}
Computing the partition function in Equation ~\ref{equ:seq_loss}
requires summing over all possible output sequences,
which is practically impossible.
Given a model,
we can first apply decoding algorithm to get $k$ high-score hypotheses $\mathbf{h}=\{\mathbf{h}_i\}_{i=1}^k$,
then approximate the partition function with $\mathbf{h}$:
\begin{equation} \label{equ:partition_approx}
\begin{aligned}
    L_{seq} & = -\log \frac{\exp(s(\mathbf{y}, \mathbf{x}))}{\sum_{\text{all}\ \mathbf{y'}}\exp(s(\mathbf{y'}, \mathbf{x}))} \\
    & \approx -\log \frac{\exp(s(\mathbf{y}, \mathbf{x}))}{\sum_{\mathbf{y'} \in \{\mathbf{y}\} \cup \mathbf{h}}\exp(s(\mathbf{y'}, \mathbf{x}))}
\end{aligned}
\end{equation}
This is equivalent to a $(k+1)$-class classification problem.
We explore two decoding algorithms to get hypotheses set $\mathbf{h}$:
standard beam search and diverse beam search ~\cite{Vijayakumar2016DiverseBS}.
Standard beam search is widely used but often produces highly similar hypotheses.
Diverse beam search can more effectively explore the search space,
and produces diverse hypotheses.

The aforementioned estimation of the partition function is biased,
in the sense that it is a strict lower bound of the actual value,
and is likely to be a very loose bound.
Noise contrastive estimation (NCE) ~\cite{Gutmann2010NoisecontrastiveEA, Deng2020ResidualEM}
provides an unbiased gradient estimator,
but the variance is expected to be high.
We leave further investigation of NCE as future work.

\subsection{Quantify Label Bias} \label{section:quantify_bias}
In this section,
we present a simple ranking-based automatic evaluation metric.
Intuitively,
a model with less label bias should give a higher score to the groundtruth text,
and lower scores to generic texts.
A piece of text is ``generic'' if it has low perplexity.
We evaluate the perplexity of all the sentences from ConvAI2 training set with GPT2
and use the top $50$ sentences as distractors.
Some examples are shown in Table ~\ref{tab:low_ppl}.

\begin{table}[ht]
\centering
\scalebox{0.9}{\begin{tabular}{c|c}
\hline
Text                         & ppl  \\ \hline
What do you want to be when you grow up? &   9.94   \\
Can you tell me a little about yourself? &   11.98   \\
What do you do for a living? &   12.22   \\
I do not know what to say to you. &  12.24  \\
What do you do in your spare time? & 13.95 \\ \hline
\end{tabular}}
\caption{Example sentences with low perplexity under pre-trained GPT2-117M language model.}
\label{tab:low_ppl}
\end{table}

Given a trained model,
the groundtruth together with these $50$ distractors are ranked
based on model-predicted scores in descending order.
We use mean rank as a metric to measure label bias.
Since distractors are selected without any prior knowledge about the input context,
they are unlikely to be appropriate outputs.
In the following sections,
experiments will show that seq2seq models completely fail at this simple ranking task.

\section{Experiments}

\begin{table*}[ht]
\centering
\begin{tabular}{c|c|cc|ccc}
\hline
Method   & Decoding   & ppl$\downarrow$ & BLEU$\uparrow$ & distinct-1(\%)$\uparrow$ & distinct-2(\%)$\uparrow$ & distinct-3(\%)$\uparrow$ \\ \hline
MLE & BS         &   \textbf{19.69}  &  3.91    &   1.20  &  5.54   &  10.62 \\
MLE & diverse BS &  \textbf{19.69}   &   \textbf{3.97}   &   1.20 &   5.87   &  11.86  \\ \hline
LogProb Avg  &   BS   &   21.73  &   3.12   &  1.89 &  10.84 &  22.32 \\
LogProb Avg &    diverse BS  &  21.06   &  3.16    &  1.87 &   11.03 &  23.37 \\
Logits Avg   &   BS    &  20.72   &  2.43    &   1.90 &   12.27 &  26.85  \\
Logits Avg  & diverse BS & 20.16 & 2.43 &  \textbf{1.91} & \textbf{12.48} & \textbf{27.98} \\ \hline \hline
Human    &     - &  -   &  -    & 3.66 &  28.89  &  60.67 \\ \hline
\end{tabular}
\caption{Automatic evaluation results.
``BS'' is short for ``beam search''.
``MLE'' uses token-level training as stated in Section ~\ref{section:token_loss}.
``LogProb Avg'' uses average log-probability as the score for sequence-level training,
while ``Logits Avg'' uses average logits as the score (Equation ~\ref{equ:logits_score}).}
\label{tab:main_results}
\end{table*}

\subsection{Setup} \label{section:setup}
\noindent
\textbf{Datasets }
We use Reddit dataset ~\cite{Dziri2018AugmentingNR} \footnote{Available at \url{https://github.com/nouhadziri/THRED}}
to pre-train our models.
It consists of more than $10$ million dialogue turns.
Reddit dataset is noisy and contains some offensive languages.
ConvAI2 dataset ~\cite{Dinan2019TheSC} \footnote{\url{http://convai.io/}} is of high quality,
and used for model fine-tuning.
To make the task more open-ended,
we discard the persona information in the ConvAI2 dataset.
Official dataset split is adopted,
$123k$ dialogue turns in the training set
and nearly $15k$ turns in the validation set.
\newline

\noindent
\textbf{Model Configuration }
Our seq2seq network uses \emph{bert-base-uncased} ~\cite{Devlin2019BERTPO} as encoder,
decoder consists of $3$ layers of Transformer blocks with hidden size $1024$.
We tie the parameters of encoder word embeddings, decoder word embeddings,
and the output softmax layer.
Adam optimizer is used with learning rate $2 \times 10^{-5}$
and batch size $32$.
We linearly warmup the learning rates in the first $4000$ updates.
The vocabulary is the same as BERT.
Dropout of $0.1$ is applied for self-attention layers,
feedforward layers,
and input embedding layers.
Gradients are clipped to have a maximum L2 norm of $1$.
We set beam size to $6$ for both hypotheses generation
in sequence-level training and model inference.
Diverse beam search uses $3$ groups.
The input context is the concatenation of the last $2$ dialogue turns.
When pre-training on Reddit,
all parameters are updated.
When fine-tuning on ConvAI2,
only decoder parameters are updated to reduce over-fitting.
Our implementation is based on \emph{fairseq}\footnote{\url{https://github.com/pytorch/fairseq}}.
\newline

\noindent
\textbf{Evaluation }
We use both automatic evaluation metrics and human evaluation
to get a comprehensive view.
Automatic evaluation metrics include perplexity (ppl),
BLEU-4 ~\cite{Papineni2001BleuAM},
and distinct-n (n=1,2,3) ~\cite{Li2016ADO}.
Distinct-n is a measure of diversity,
it computes the number of distinct n-grams,
normalized by the number of all n-grams.
Mean rank among pre-defined distractors is also reported,
as stated in Section ~\ref{section:quantify_bias}.
For human evaluation,
we use the Specificity and Sensibleness Average (SSA)
proposed by Meena ~\cite{Adiwardana2020TowardsAH}.
Two annotators are asked to score each response
for $200$ randomly chosen dialogue turns from the validation set
on a scale of $0$ to $2$ ($0$-bad, $1$-ok, $2$-good).
SSA score is the arithmetic mean of specificity score and sensibleness score,
it measures the quality of generated texts
based on two complementary dimensions.

\subsection{Results}

\begin{table*}[ht]
\centering
\scalebox{1.0}{\begin{tabular}{c|l}
\hline
Input      & \textbf{A:} Hello what are you doing today? \\ \hline
MLE        & I'm doing well. How are you? \\
Logits Avg & Hi! I'm doing some shopping and enjoying a good steak! you? \\
Human & I am good, I just got off work and tired. I have two jobs. \\ \hline \hline
Input      & \begin{tabular}[c]{@{}l@{}} \textbf{A:} I have a turtle his name is speedy, kitties are nice too, tho! \\ \textbf{B:} That is an adorable name for a turtle. I have 2 cats. \end{tabular} \\ \hline
MLE        & What do you do for a living? \\
Logits Avg      & They're so cute! Do you have a favorite band? \\
Human &  What are your kitties names? \\ \hline \hline
Input      & \begin{tabular}[c]{@{}l@{}} \textbf{A:} That's okay we all have our differences in likes. \\ \textbf{B:} That is very true \end{tabular} \\ \hline
MLE        & What do you do for a living? \\
Logits Avg      & What do you do for a living? \\
Human & I write poetry and then make them into a song. \\ \hline \hline
Input      & \begin{tabular}[c]{@{}l@{}} \textbf{A:} I do speak multiple languages. Which ones do you speak? \\ \textbf{B:} I speak, English, Spanish and French. \end{tabular} \\ \hline
MLE        & Do you speak any other languages? \\
Logits Avg      & What language do you speak? \\
Human & Those happen to all be languages I speak. I want to visit France sometime. \\ \hline \hline
\end{tabular}}
\caption{Examples from ConvAI2 validation dataset.
Input only lists the last $2$ dialogue turns.
Both ``MLE'' and ``Logits Avg'' use diverse beam search decoding.
Outputs from ``LogProb Avg'' are not shown due to space limit.
``\textbf{A}'' and ``\textbf{B}'' are used to denote different persons.}
\label{tab:examples}
\end{table*}

Table ~\ref{tab:main_results} shows the main results for automatic evaluation.
The token-level cross entropy loss used by ``MLE''
is specifically targeted for optimizing perplexity.
Not surprisingly,
``MLE'' achieves the lowest perplexity of $19.69$.
``MLE'' also has a higher BLEU score of $3.97$ than sequence-level training methods.
One possible explanation is that
there are many more ways to be specific than to be generic.
Producing a generic output is more likely to match the groundtruth,
and get a higher BLEU score.
Though BLEU is widely used in evaluating machine translation systems,
previous work ~\cite{Liu2016HowNT} suggests that
BLEU only has a weak correlation with human evaluation results for response generation.

Distinct-n metric measures the diversity of generated texts.
Based on distinct-n (n=1,2,3) in Table ~\ref{tab:main_results},
both sequence-level training methods ``LogProb Avg'' and ``Logits Avg''
produce significantly more diverse results than baseline ``MLE'' methods.
In terms of decoding algorithm,
diverse beam search shows consistent improvements across nearly all automatic metrics.

``LogProb Avg'' uses average log-probability as score,
token-level loss and sequence-level loss may compete for the same probability mass.
Perplexity increases from $19.69$ to $21.06$.
``Logits Avg'' can calibrate the logits
while keeping the local probability relatively unchanged
as shown in Equation ~\ref{equ:calibrate_logits}.
Perplexity only slightly increases from $19.69$ to $20.16$.

\begin{table}[ht]
\centering
\scalebox{0.92}{\begin{tabular}{c|ccc}
\hline
          & MLE & LogProb Avg & Logits Avg \\ \hline
Mean Rank$\downarrow$ &  44.7  &   43.9 &  \textbf{35.1} \\ \hline
\end{tabular}}
\caption{Mean rank of groundtruth among $50$ context-agnostic distractors.
Lower mean rank indicates the model has less label bias.
See Section ~\ref{section:quantify_bias} for more details.}
\label{tab:rank_metric}
\end{table}

In Table ~\ref{tab:rank_metric},
``MLE'' fails miserably at discriminating groundtruth
from $50$ pre-defined distractors with a mean rank of $44.7$.
``Logits Avg'' performs best among 3 methods with a mean rank of $35.1$.
However,
a naive baseline that randomly shuffles all the candidates would have a mean rank of $25.5$,
which is far better than our best model.
This is evidence that
our proposed method only reduces label bias to some degree
instead of eliminating it.

\begin{table}[ht]
\centering
\scalebox{0.95}{\begin{tabular}{c|ccc}
\hline
Method      & Specificity & Sensibleness & SSA \\ \hline
MLE         &  0.54 &   0.88  &   0.71  \\
LogProb Avg &  0.68  &  1.00   &   0.84      \\
Logits Avg  & \textbf{1.06}   & \textbf{1.24}  &   \textbf{1.15}  \\ \hline
Human       &  1.60  &  1.47  &   1.53   \\ \hline
\end{tabular}}
\caption{Human evaluation results.
The scores are averaged over two annotators and $200$ dialogue turns,
and are in the range of $0$ to $2$.
All methods adopt diverse beam search as the decoding algorithm
since it shows slightly better performance on automatic evaluation metrics.
``SSA'' is the arithmetic mean of specificity score and sensibleness score.}
\label{tab:human_eval}
\end{table}

We conduct a human evaluation on $200$ random dialogue turns from the validation dataset.
Results are in Table ~\ref{tab:human_eval}.
Models tend to generate sensible but not very specific texts,
the sensibleness score for all $3$ models are
higher than the corresponding specificity score in Table ~\ref{tab:human_eval},
while human texts are much more specific.
``MLE'' produces generic texts with a very low specificity score of $0.54$.
Both ``LogProb Avg'' (SSA $0.84$) and ``Logits Avg'' (SSA $1.15$)
improves over the ``MLE'' baseline (SSA $0.71$),
showing sequence-level training can indeed lead to more specific and sensible outputs.
Using unnormalized logits as the score is more effective than using log-probabilities.
Also,
sequence-level training has a larger impact
on specificity ($+96\%$ relative increase from $0.54$ to $1.06$)
than sensibleness ($+41\%$ relative increase from $0.88$ to $1.24$).

\subsection{Analysis}
Some typical examples are given in Table ~\ref{tab:examples}.
The first two examples showcase that
``MLE'' often generates generic texts such as ``I'm doing well'',
``What do you do for a living?'', etc.
Many previous works also reported similar findings ~\cite{Dziri2018AugmentingNR, Adiwardana2020TowardsAH}.
Our proposed method ``Logits Avg'' can generate meaningful and specific words
such as ``enjoying a good steak'', ``favorite band'', etc.
It also illustrates why BLEU may not be a good metric to evaluate open-ended text generation systems.
Though outputs by ``Logits Avg'' are of high quality,
there are not many overlapping words with groundtruth.

The last two examples in Table ~\ref{tab:examples}
show some existing limitations and difficulties for open-ended generation.
In the third example,
both ``MLE'' and ``Logits Avg'' produce the same generic response,
another evidence that sequence-level training does not
completely solve the label bias problem in seq2seq networks.
Beam search is not guaranteed to find the optimal output sequence,
but this may be a good thing for promoting response diversity.
In the fourth example,
``Logits Avg'' asks a question that
has already been answered in previous dialogue turns.
Generating semantically consistent responses is still an open problem.

\begin{figure}[ht]
\begin{center}
 \includegraphics[width=1.0\linewidth]{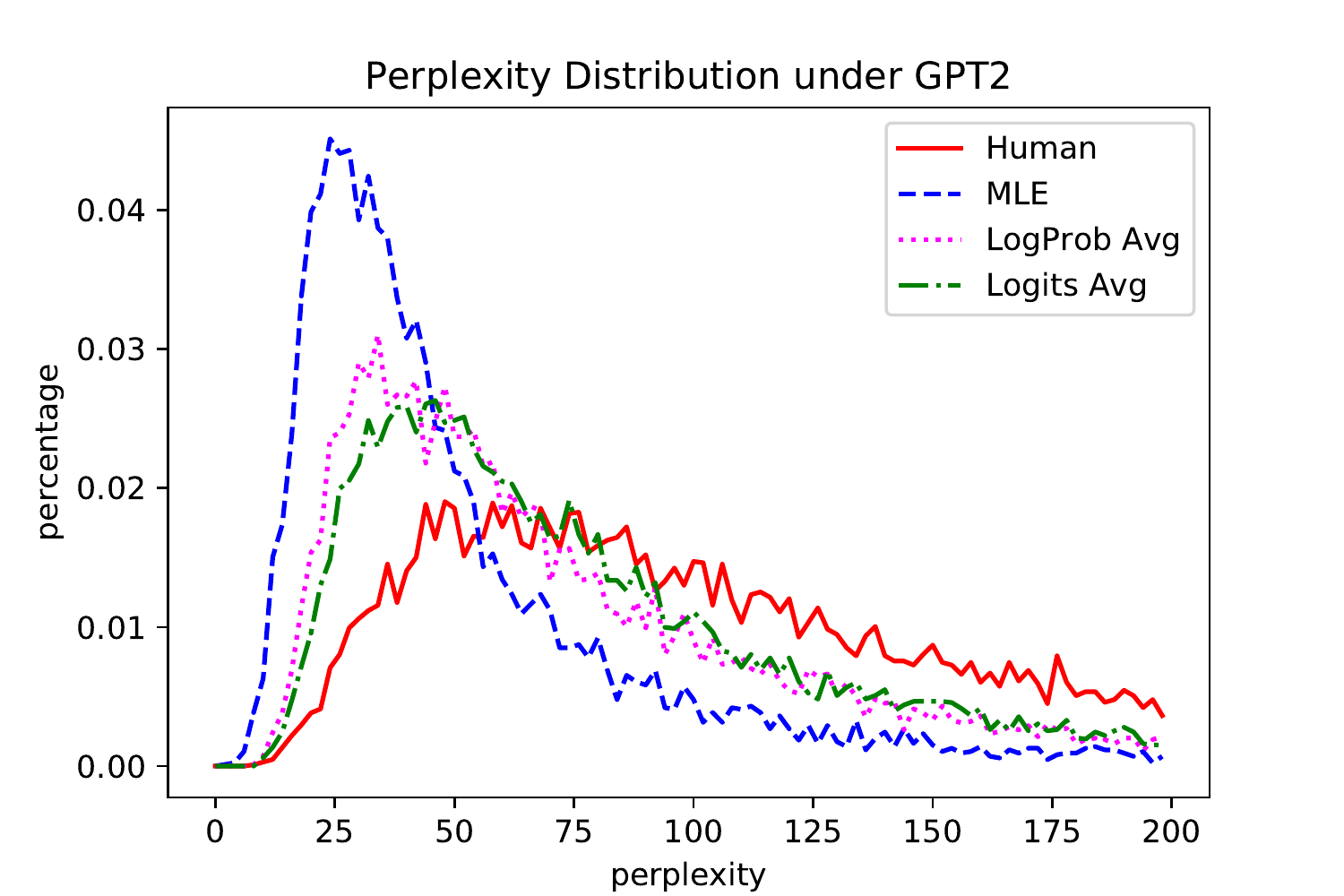}
 \caption{Perlexity distribution of texts generated by different models under GPT2-117M.
 Better viewed in color.}
 \label{fig:ppl_4gpt}
\end{center}
\end{figure}

In Figure ~\ref{fig:ppl_4gpt},
we additionally show the perplexity distribution of texts
from two sequence-level training models ``LogProb Avg'' and ``Logits Avg''.
The distributions of both models are flatter than ``MLE'' baseline,
and the peaks move to the right.
The perplexity distribution of ``Logits Avg''
is slightly closer to humans than ``LogProb Avg''.
Though our proposed methods are less biased,
they still prefer low-perplexity texts compared to humans.

\subsection{Discussion}
Label bias arises when
different output states have very different numbers of outgoing transitions.
In directed generation tasks
such as machine translation and abstractive summarization,
there is a nearly one-to-one mapping between the input and the output.
The transitions between the output states are almost deterministic,
thereby label bias exists but is not a serious issue.
Previous work ~\cite{Andor2016GloballyNT, Edunov2018ClassicalSP}
observes some moderate improvements with globally normalized training.
It remains to be seen how state-of-the-art text generation models
based on BERT and GPT are affected by label bias.

In linear-chain CRF ~\cite{Lafferty2001Conditional},
partition function can be accurately and efficiently computed with
the Viterbi algorithm based on dynamic programming.
However,
in seq2seq networks,
outputs at each timestep couples with each other,
and can not fit into the framework of the Viterbi algorithm.
In this paper,
we use beam search results to estimate the partition function.
Such inaccuracy may be one major reason why
our proposed model still favors generic texts to a large degree.

In token-level MLE training,
each update requires one forward pass and one backward pass.
In sequence-level training,
an extra decoding step is required.
Auto-regressive decoding is a sequential process,
and therefore is pretty slow.
It prevents us from fully exploiting the computation power of modern GPUs
and the inherent parallelizability of Transformers.
Common practices ~\cite{Edunov2018ClassicalSP} first pre-train the network with token-level MLE,
and then finetune with sequence-level loss.

\section{Related Work}
\textbf{Neural Text Generation }
with seq2seq models has been a popular paradigm for many generation tasks in recent years,
such as neural machine translation ~\cite{Wu2016GooglesNM},
abstractive summarization ~\cite{See2017GetTT},
and grammatical error correction ~\cite{Zhao2019ImprovingGE}, etc.
Most existing models use token-level maximum likelihood estimation as optimization objective,
and beam search as sequence decoding algorithm.
The backbone architecture includes LSTM,
CNN ~\cite{Gehring2017ConvolutionalST} and Transformers ~\cite{Vaswani2017AttentionIA}.
Since Transformers are highly parallelizable
and have the ability to model long-term dependencies,
they have become a core component for many state-of-the-art models ~\cite{Radford2018ImprovingLU}.
Exposure bias ~\cite{Bengio2015ScheduledSF, Zhang2019BridgingTG}
is widely studied in seq2seq models trained with teacher forcing.
With the emergence of various powerful pre-trained models
like BERT ~\cite{Devlin2019BERTPO} and GPT-2 ~\cite{Radford2019LanguageMA},
there are growing interests in improving text generation
with language model pre-training ~\cite{Song2019MASSMS, Wang2019DenoisingBS}.
\newline

\noindent
\textbf{Beam Search }
with length normalization is a widely used heuristic sequence decoding algorithm
for many structured prediction models ~\cite{Wu2016GooglesNM, Bahdanau2015NeuralMT}.
It has several known deficiencies,
including length bias ~\cite{Yang2018BreakingTB, Huang2017WhenTF},
lack of diversity within beams ~\cite{Vijayakumar2016DiverseBS},
and performance degradation with larger beams ~\cite{Cohen2019EmpiricalAO} ~\cite{Stahlberg2019OnNS}, etc.
In open-ended text generation such as story generation ~\cite{Fan2018HierarchicalNS},
conditional language modeling ~\cite{Holtzman2019TheCC},
standard beam search is found to often produce degenerate outputs
and therefore are rarely used.
In sampling-based decoding algorithms,
tricks like adjusting temperature and
explicitly blocking duplicate n-grams work well ~\cite{Fan2018HierarchicalNS}.
Some heuristic methods
are proposed to promote the diversity of beam search outputs.
~\citeauthor{Xu2019NeuralRG} incorporate additional meta-words into the context,
~\citeauthor{Gao2019JointlyOD} jointly optimize both diversity and relevance with variational auto-encoders,
and ~\citeauthor{Li2016ADO} rerank beam search outputs based on Maximum Mutual Information (MMI).
\newline

\noindent
\textbf{Label Bias }
is usually associated with locally normalized models for structured prediction,
such as Maximum Entropy Markov Models (MEMM) ~\cite{McCallum2000MaximumEM}.
Label bias ~\cite{Hannun2020TheLB} makes the model prefer states with fewer outgoing transitions
and makes it difficult to correct past mistakes.
Conditional random field (CRF) ~\cite{Lafferty2001Conditional, Koller2009ProbabilisticGM}
eliminates label bias with global normalization.
More generally,
undirected graphical models ~\cite{Koller2009ProbabilisticGM} do not suffer label bias
like most directed graphical models do.
However,
computing the partition function can be difficult
without strong conditional independence assumptions.
Sequence-level training approximates the partition function
with decoded hypotheses ~\cite{Andor2016GloballyNT, Collobert2019AFD},
and proves to be effective in neural machine translation ~\cite{Edunov2018ClassicalSP},
part-of-speech tagging ~\cite{Andor2016GloballyNT, Le2013OnTE},
speech recognition ~\cite{Collobert2019AFD},
and summarization ~\cite{Wiseman2016SequencetoSequenceLA}, etc.
~\citeauthor{Deng2020ResidualEM} adopt noise contrastive estimation to
train residual energy models for text generation.
Yet little attention has been paid to the effect of label bias for
seq2seq models in open-ended text generation scenarios.

\section{Conclusion and Future Work}
The degenerate behaviors of beam search in open-ended generation have been long recognized.
This paper empirically investigates the effects of label bias for beam search
based on the response generation task.
Likelihood distribution evaluation shows beam search outputs are biased
towards low-perplexity generic texts,
and this phenomenon is mostly attributed to model errors.
Globally-normalized sequence-level training can help reduce label bias.
Using logits as scores is more effective than using log-probabilities.
We also propose a simple ranking-based metric to measure label bias.
Experiments show beam search can produce more diverse outputs with our proposed method.
Due to the difficulty of estimating partition function,
more research efforts are still needed to eliminate label bias.

For future work,
we would like to investigate label bias in
other open-ended generation tasks like conditional language modeling,
and story generation.
Another important research direction
is to explore more effective and
efficient methods for globally normalized training.


\bibliography{aacl-ijcnlp2020}
\bibliographystyle{acl_natbib}

\end{document}